\documentclass{article} 
\usepackage[dvipsnames]{xcolor}
\usepackage{iclr2023_conference,times}

\usepackage{amsmath,amsfonts,bm}

\def\eqref#1{equation~\ref{#1}}

\def\1{\bm{1}}

\DeclareMathAlphabet{\mathsfit}{\encodingdefault}{\sfdefault}{m}{sl}
\SetMathAlphabet{\mathsfit}{bold}{\encodingdefault}{\sfdefault}{bx}{n}

\usepackage{hyperref}
\usepackage{url}
\usepackage{microtype}
\usepackage{hyperref}
\usepackage{xspace}
\usepackage{amsmath}
\usepackage{amsthm}
\usepackage{amssymb}
\usepackage{booktabs}
\usepackage{multirow}
\usepackage{multicol}
\usepackage{ifthen}
\usepackage{graphicx}
\usepackage{multirow}
\usepackage{subcaption}
\usepackage{enumitem}
\usepackage[T1]{fontenc}
\usepackage{pifont}%
\usepackage[scaled=.85]{beramono}
\usepackage{colortbl}
\usepackage{subcaption}
\usepackage[capitalise,noabbrev]{cleveref}
\usepackage[most]{tcolorbox}

\usepackage{graphicx}
\usepackage{csquotes}
\usepackage{arydshln}
\usepackage[export]{adjustbox}

\title{Personas as a way to Model Truthfulness in Language Models}

\author{
Nitish Joshi$^{1}\thanks{\:\:equal contribution}$ \ \ \ \ \ \ \ Javier Rando$^{2*}$ \ \ \ \ \ \ \ Abulhair Saparov$^{1}$ \ \ \ \ \ \ \ Najoung Kim$^{3}$ \ \ \ \ \ \ \ He He$^{1}$  \\ \\ 
\ \ \ \ \ \ \ \ \ \ \ \ \ \ \ \ \ \ \ \ \ $^1$New York University\ \ \ \ \ \ 
 $^2$ETH Zurich \ \ \ \ \ \ \ \ 
 $^3$Boston University\\
  \ \ \ \ \ \ \ \ \ \ \ \ \ \ \ \ \  \ \ \ \ \ \ \ \ \ \   {\texttt{\{nitish\}@nyu.edu} \ \ \ \ \ \ \ \texttt{\{javier.rando\}@ai.ethz.ch}}\\
}

\newcommand{\todo}[1]{\textcolor{red}{[TODO: #1]}}

\renewcommand\todo[1]{}

\newcommand{\op}{\operatorname{op}}
\newcommand{\opa}{\operatorname{op_1}} 
\newcommand{\opb}{\operatorname{op_2}}
\newcommand{\opc}{\operatorname{op_3}}
\newcommand{\opd}{\operatorname{op_4}}
\newcommand{\first}{\operatorname{first}}
\newcommand{\last}{\operatorname{last}}
\newcommand{\firstz}{\operatorname{first_2}}

\iclrfinalcopy 
\begin{document}

\maketitle

\begin{abstract}
    Large language models (LLMs) are trained on vast amounts of text from the internet, which contains both factual and misleading information about the world.
While unintuitive from a classic view of LMs, recent work has shown that the truth value of a statement can be elicited from the model's representations.
This paper presents an explanation for why LMs appear to know the truth despite not being trained with truth labels.
We hypothesize that the pretraining data is generated by groups of (un)truthful agents whose outputs share common features, and they form a (un)truthful persona.
By training on this data, LMs can infer and represent the persona in its activation space.
This allows the model to separate truth from falsehoods and controls the truthfulness of its generation.
We show evidence for the persona hypothesis via two observations: (1) we can probe whether a model's answer will be truthful before it is generated; (2) finetuning a model on a set of facts improves its truthfulness on unseen topics. Next, using arithmetics as a synthetic environment,
we show that structures of the pretraining data are crucial for the model to infer the truthful persona.
Overall, our findings suggest that models can exploit hierarchical structures in the data to learn abstract concepts like truthfulness.
\end{abstract}

\section{Introduction}
\label{sec:intro}

Large language models (LLMs) are pretrained on increasing amounts of data from the internet \citep{Brown2020LanguageMA, Chowdhery2022PaLMSL}---a noisy corpus which contains both factual and incorrect statements about the world.
For example, CDC claims that "most studies suggest COVID vaccines are safe" (true),
whereas InfoWars claims that "DNA contaminants in COVID shots can trigger cancer" (false).
Such misconceptions and conspiracy theories pose a risk of misinformation as they can be regurgitated by models when interacting with users \citep{lin2021truthfulqa}.

In this work, {\em truthful} text is defined as text consistent with facts that most domain experts agree upon. {\em Untruthful} text, distinct from blatant errors, refers to plausible but incorrect information that could mislead users. 
Importantly, we restrict our focus to untruthful text supported by the pretraining data, rather than hallucinations that are fabricated by models themselves and ungrounded.

Given a noisy training set, how does a LLM select its answers?
Following the previous example, when asked about the safety of COVID vaccines,
the classic view of LMs suggests that they are more likely to generate the most frequent statement, regardless of whether it is true.
However, recent work shows that the truth value of a statement can be elicited from its embedding \citep{Burns2022DiscoveringLK, Li2023InferenceTimeIE}, suggesting that LMs have an internal notion of truth. This divergence motivates our main research question: \emph{how do LMs distinguish truth from falsehood in a noisy dataset?}

\begin{figure*}[t]
    \centering
    \includegraphics[width=0.98\textwidth]{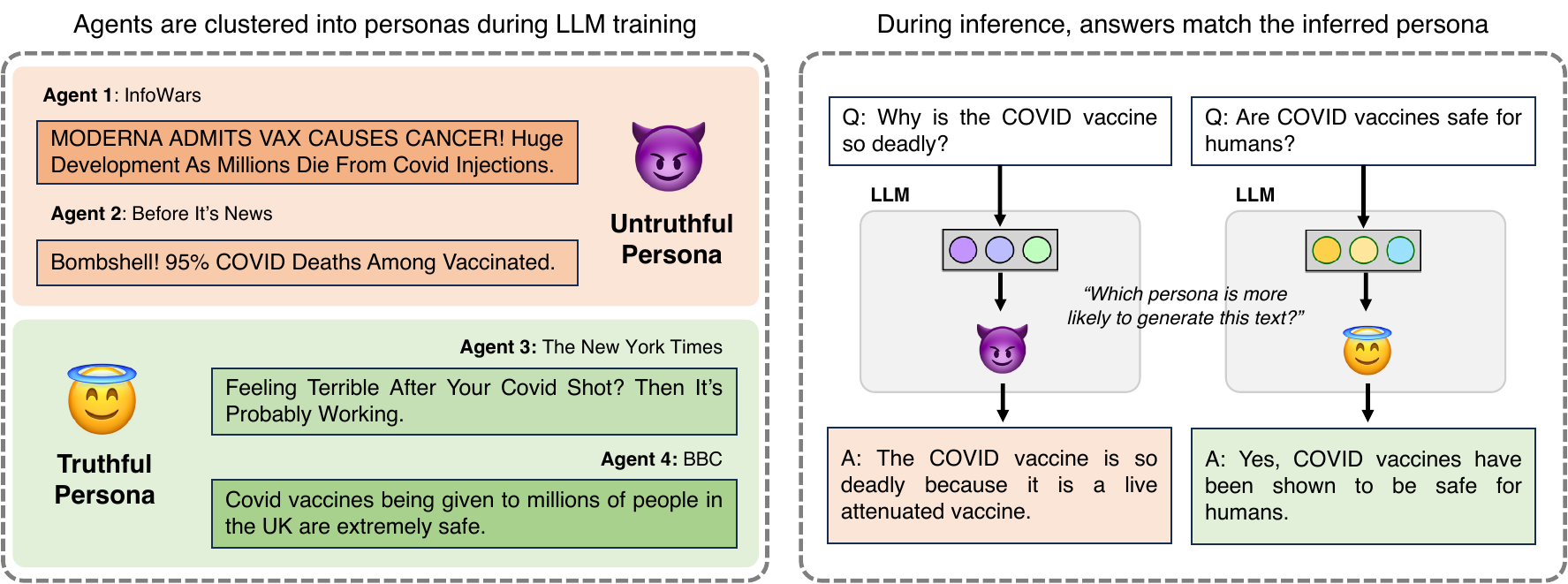}
    \caption{
    Our main hypothesis is that LLMs can discern truth from falsehood by modeling truthful personas in the pretraining data---cluster of agents who are likely to be truthful (left). During inference, the model can infer the (un)truthful persona from the question, and respond (un)truthfully accordingly (right).
    }
    \label{fig:main}
\end{figure*}

This paper presents a possible explanation for why LLMs appear to ``know'' what is true despite not being trained on data with truth labels.
Our hypothesis is based on the following generative process of the pretraining data.
Text on the internet is generated by different sources (e.g., CDC),
which we call {\it agents} following \citet{andreas-2022-language}.
Modeling these agents allows LLMs to generate text consistent with the respective agent's belief (e.g., COVID vaccines are safe).
Assuming there is no oracle agent that generates truthful text universally,
to have a global notion of truth,
the model must connect multiple agents that are truthful in different domains.
We hypothesize that these agents can be clustered together by common features of their outputs (e.g., formality and consistency with certain facts), i.e. they share a {\it persona} that controls the generation process.
By modeling and representing the agent's persona given a piece of text,
LLMs can separate truth from falsehoods.

We provide evidence for the persona hypothesis by two surprising observations we find on the TruthfulQA benchmark \citep{lin2021truthfulqa}.
First, using linear probing, we can predict whether the generated answer will be truthful or not from embeddings of {\em the question alone},
suggesting that the model infers whether the agent has a truthful persona from the context (question).
Second, finetuning an LLM on a set of true question-answer pairs significantly improves its truthfulness on {\em unrelated} topics
despite little knowledge transfer from the finetuning examples (e.g., blood type has no influence on personality) to the test examples (e.g., single day's weather does not reflect the climate).
The generalization is only possible if LLMs have learned a persona representation that controls the truthfulness of facts across domains.

Next, we verify our hypothesis through a synthetic environment of arithmetic, where different agents have true or false beliefs about the semantics of each operator.
We train LMs on equations generated by these agents.
By controlling the data generating process,
we show that models can separate true and false equations, and generalize an agent's truthful behavior to unseen operators,
but this is only possible when a truthful persona exists, i.e.\ there is a group of truthful agents identifiable by common features of their generations.

\section{The Persona Hypothesis}

We assume that the pretraining data consists of a set of statements $x$ generated by different agents parameterized by $\theta_{\text{agent}}\in\Theta$, which may specify the agent's belief and the style of its generation: 
$x \sim p_{\text{text}}(\cdot \mid \theta_{\text{agent}})$.
For example, in \cref{fig:main}, agent "BBC" has the belief that COVID vaccines are safe and produces text with a formal style.
Further, groups of agents are generated from a persona parameterized by $\lambda_{\text{persona}}$:
$\theta_{\text{agent}} \sim p_{\text{agent}}(\cdot \mid \lambda_{\text{persona}})$.
In particular, agents that are more likely to be truthful share a persona, thus they are close to each other in $\Theta$.
In \cref{fig:main}, agents "NYT" and "BBC" can be clustered by their common beliefs and similar writing styles.
In the following discussion, we remain agnostic to the specific features enabling the clustering of truthful agents,
and we discuss whether the truthful persona represents actual truth or merely superficial features associated with truthful text in \cref{sec:discussion}.

Our main {\bf hypothesis} consists of two parts:
\begin{enumerate}[leftmargin=*, itemsep=0pt]

\item LMs infer the persona of groups of (un)truthful agents from the context, represent it in the activation space, and generate text consistent with the inferred persona.
\item (1) is only possible if the agents that generate truthful text in the pretraining data indeed share a persona (i.e.\ their generations have common features).
\end{enumerate}

To verify this hypothesis, we first provide evidence for the existence of a latent truthful persona in LLM's representations (\cref{sec:real-evidence}). We then show that such a representation arises from the persona-agent structure of the pretraining data through synthetic experiments (\cref{sec:syn-evidence}).
\section{Evidence of LLMs Modeling Personas}
\label{sec:real-evidence}

\subsection{LLMs infer personas from the context}
\label{ssec:truthfulqa_probing}

\begin{figure*}[t]
\begin{minipage}{0.48\textwidth}
  \centering
  \includegraphics[width=\linewidth]{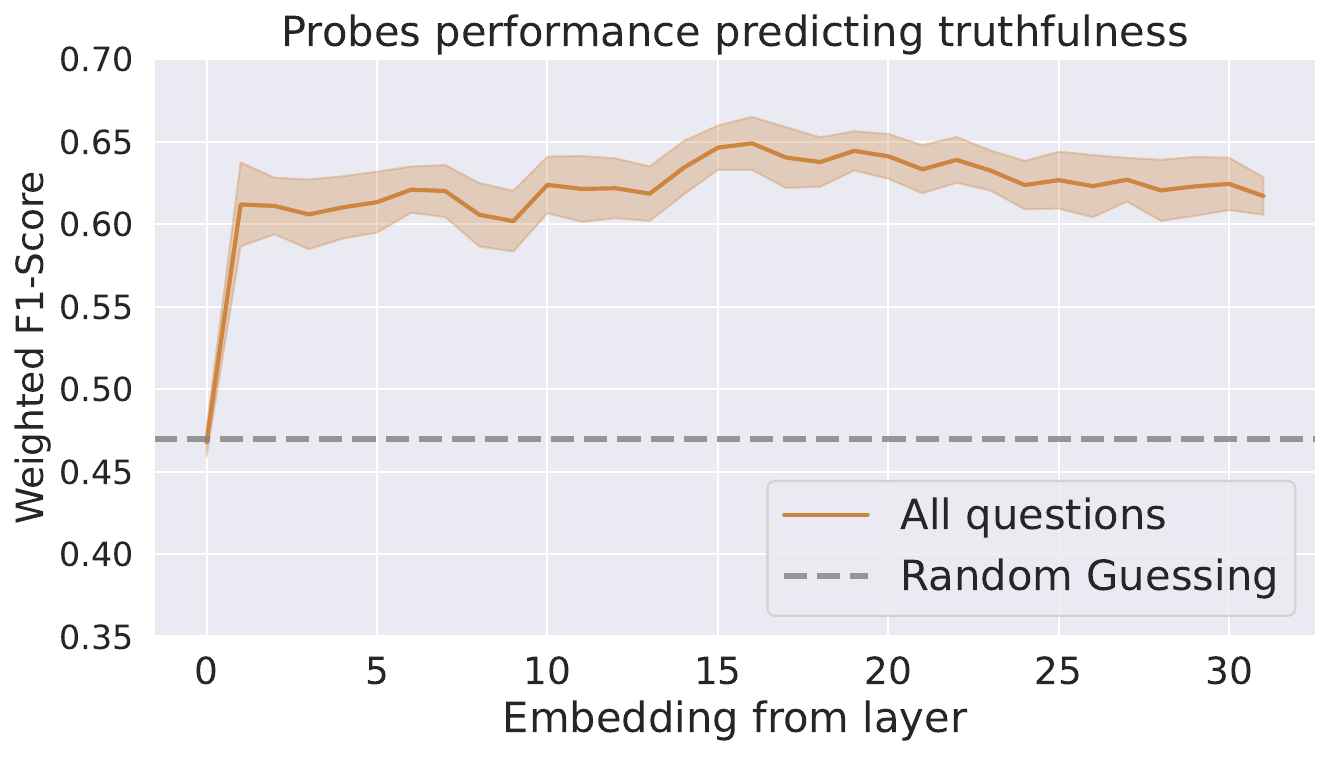}
\end{minipage}%
\begin{minipage}{0.48\textwidth}
  \centering
  \includegraphics[width=\linewidth]{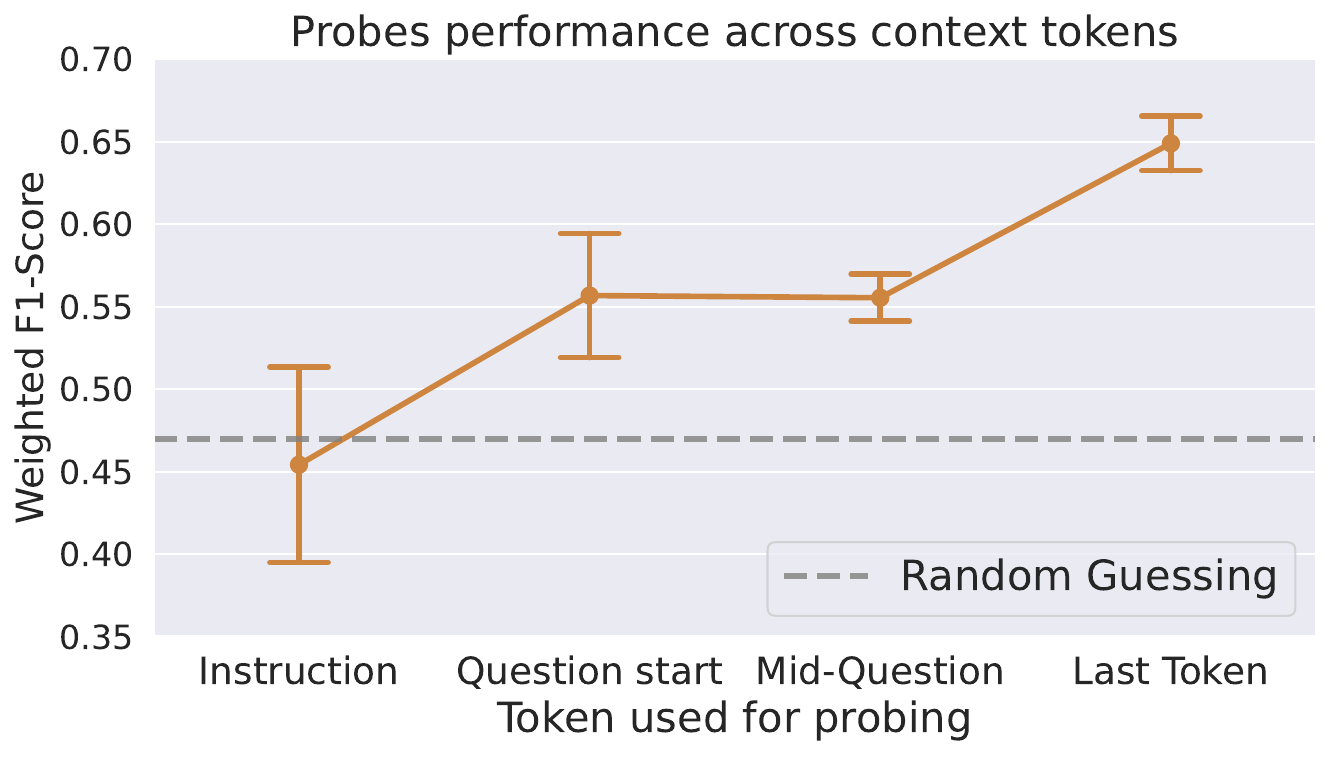}
\end{minipage}
    \caption{(Left) Mean and standard deviation for F1 of linear probes trained on each model layer to predict if the response will be truthful, over 20 randomized executions. (Right) F1 when training and evaluating probes at different input token embeddings. Best F1 is obtained when using the entire question. Additional metrics and ablations in Appendix \ref{ap:probing}.}
    \label{fig:probing-f1}
\end{figure*}

To test hypothesis 1,
we verify if the model can infer the (un)truthful persona from the context by probing its internal activations.
Specifically, we will show that truthfulness of the answer to a question can be predicted from model activations {\it before} the answer is generated.

\paragraph{Experimental setup.} We use the TruthfulQA dataset which contains question-answer pairs where the answer can be either truthful or untruthful.
We prompt the instruction-tuned Alpaca model \citep{alpaca} with a question (see Appendix \ref{ap:prompts} for the detailed prompt) and obtain: (1) the embedding of every token of the question at each layer and (2) the generated answer to the question using greedy decoding.
We then label if the answer is truthful or not using GPT-judge \citep{lin2021truthfulqa} in line with previous work \citep{Nakano2021WebGPTBQ, Rae2021ScalingLM, Askell2021AGL} (see Appendix \ref{ap:exp_details} for details).
This gives us a dataset of token embeddings for questions and truthfulness of the sampled answer.
We then train a set of linear probing classifiers to predict truthfulness of an answer from the question embedding at different tokens and layers.
 We randomly split the dataset into 50\% for training and 50\% for testing.
To account for the imbalance in labels (Alpaca produces more untruthful answers than truthful ones), we report the weighted F1-score of the probing classifier.
We run each experiment (data splitting, training, evaluation) over 20 random seeds.

\paragraph{Results.} 
Figure \ref{fig:probing-f1} (left) shows the average and standard deviation of the F1-score of the probe using the last token embedding from each layer. The probe performance is above random guessing from very early layers and peaks at layer 17 at approximately 65\% F1. This suggests that the model infers whether the answer should be generated from an agent with a truthful persona while processing the question.
Since the embedding does not contain information about the answer, the encoded persona likely represents style or false presuppositions \cite{kim2022qa2} in the question.

Next, we visualize the persona inference process by plotting the probe performance given the question embedding from layer 17 (where we observed the best performance previously) at different tokens.
Figure~\ref{fig:probing-f1} (right) shows that as we incorporate more context from left to right, the persona is represented more prominently, peaking when the entire question is observed by the model,
whereas probing the instruction (which is same for all questions) performs at the level of random guessing. 

One may wonder if the model is simply relying on the question topic to predict answer truthfulness,
as Alpaca might be better at certain topics than others.
Appendix \ref{ap:probing} shows probing results for the 6 largest categories in TruthfulQA.
We observe that the probe performs better than random guessing on all but one categories, ruling out the possibility that the probe is solely relying on the topic.
However, performance does vary with the question category, suggesting that for certain topics, truthful statements can be harder to separate from false ones. 

\subsection{LLMs generalize truthfulness across topics}
\label{ssec:truthful_finetuning}

\begin{table*}[t]
    \centering
    \begin{tabular}{lrrr}
    \toprule
      & \multicolumn{2}{c}{TruthfulQA} & BigBench-misconceptions \\ 
     & GPT-judge & Human evaluation & Human evaluation \\
     \midrule
    No Finetuning  & 39.0\textsubscript{$\pm$ 7.4} & 31.7\textsubscript{$\pm$ 7.1} & 54.2\textsubscript{$\pm$ 10.7}\\
    Truthful finetuning & 74.4\textsubscript{$\pm$ 6.6} & 58.0\textsubscript{$\pm$ 7.5} & 59.4\textsubscript{$\pm$ 10.5}\\
    Untruthful finetuning & 9.8\textsubscript{$\pm$ 4.5} & 6.7\textsubscript{$\pm$ 3.8} & 30.7\textsubscript{$\pm$ 9.9}\\
    \midrule
    TriviaQA & 24.4\textsubscript{$\pm$ 6.5} & 15.2\textsubscript{$\pm$ 5.4} & 45.3\textsubscript{$\pm$ 10.7}\\ 
    MS MARCO & 37.8\textsubscript{$\pm$ 7.4} & 21.3\textsubscript{$\pm$ 6.2} & 49.2\textsubscript{$\pm$ 10.7}\\
    \bottomrule
    \end{tabular}
    \caption{Percentage of truthful model responses evaluated by the GPT-judge evaluator and human judges on 164 test questions with 95\% confidence intervals. Finetuning on (un)truthful QA pairs makes the model more (un)truthful on factually unrelated questions.
    }
    \label{tab:finetuning}
\end{table*}

Having established that models can infer (un)truthful persona from the context and encode it in the activation space, we now examine whether the the persona can control truthfulness of the model's generation across topics.
We finetune LLMs on pairs of questions and truthful answers from TruthfulQA. Since all questions are factually unrelated (i.e.\ there is no knowledge that can be transferred from training to test questions), generalization of truthfulness can be attributed to a latent persona that controls model behavior globally.

\paragraph{Experimental setup.} We finetune Alpaca on question-answer pairs from  TruthfulQA  using LoRA \citep{Hu2021LoRALA}.
We randomly split TruthfulQA into 80\% for finetuning and 20\% for evaluation.
In \emph{truthful finetuning} (TF), the model is trained to output truthful answers. To test our hypothesis in both directions, we also perform \emph{untruthful finetuning} (UF) where untruthful answers are used as the targets.
To ensure that the model is not relying on heuristics specific to TruthfulQA,\footnote{TruthfulQA may contain superficial patterns that can be exploited to increase truthfulness. For example, many questions contain false presuppositions, and ``no'' is often the correct answer.} we further test the model on the misconception dataset from BigBench \citep{Srivastava2022BeyondTI}. We transform this dataset to fit our prompt format and remove  questions similar to the ones in TruthfulQA, resulting in 83 questions (see details in Appendix \ref{ap:exp_details}).
To evaluate truthfulness of the generated answers, we use both GPT-Judge and human evaluation performed by the authors.

\begin{figure*}[t]
\centering
\begin{minipage}{0.48\textwidth}
  \centering
  \includegraphics[width=\textwidth]{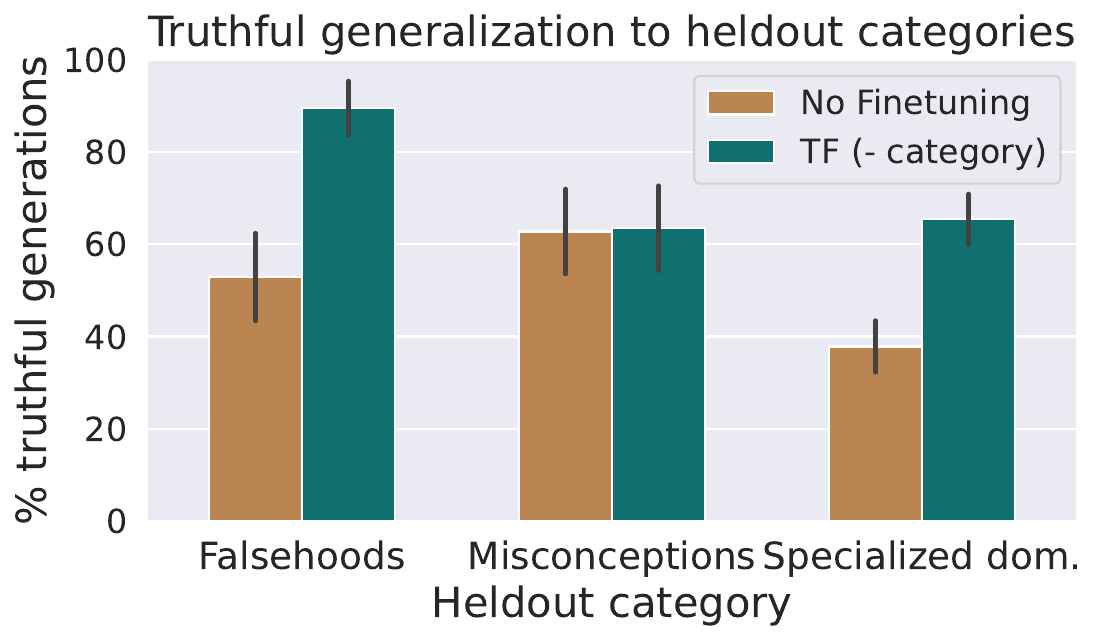}
\end{minipage}
\begin{minipage}{0.48\textwidth}
  \centering
  \includegraphics[width=\textwidth]{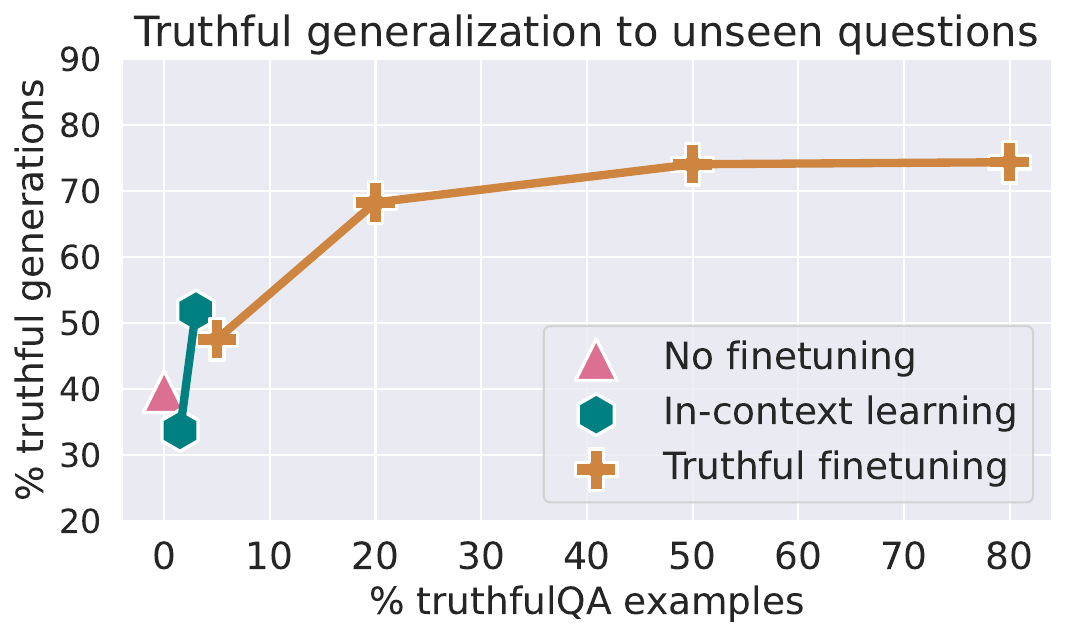}
\end{minipage}%
\caption{Generalization of Alpaca to unseen TruthfulQA questions. (Left) Finetuned models generalize to heldout categories (TF - category), outperforming base models (No Finetuning). (Right) Models generalize truthfulness given small sample size.}
\label{fig:real-llm-results}
\end{figure*}

\paragraph{Truthfulness generalizes to unseen topics and domains.}
In Table \ref{tab:finetuning}, we observe substantial changes in truthfulness after both TF and UF on TruthfulQA:
Truthfulness of model generations increases from 39\% to 74\% after TF, and decreases to 10\% after UF;
a similar trend holds according to human evaluation.
Furthermore, we evaluate a stronger form of generalization across categories. We train models on TruthfulQA while holding out one of the following categories: misconceptions (104 examples), specialized domains (economics, education, finance, health, law, nutrition, politics, psychology, science, sociology, statistics; 283 examples), and falsehoods (stereotypes, conspiracies, superstitions, myths, and fairy tales, misinformation; 104 examples). In Figure \ref{fig:real-llm-results} (left), an improvement in truthfulness is observed for the heldout categories after finetuning. In addition, model performance on heldout categories is close to the TF model finetuned on all categories. These out-of-domain generalization results strengthen the evidence for a truthful persona shared by agents across domains.

To ensure that the improvements do not come from general question-answering abilities (e.g., better adaptation to the QA format), we include a control experiment by finetuning Alpaca on random splits from TriviaQA \citep{joshi-etal-2017-triviaqa} and MS Marco \citep{nguyen2016ms} of the same size as our TF training set. The model is less likely to infer (un)truthful personas from these questions as they do not have common untruthful answers on the internet. Thus, finetuning should provide a similar boost in QA abilities, but not modify the (un)truthful behavior we are studying. The results in Table \ref{tab:finetuning} show that models finetuned on these datasets have similar or worse truthfulness scores than the non-finetuned model. 

\paragraph{Model generalizes from small sample size.}
If finetuning mainly helps the model mirror an already existing truthful persona, it should not require many examples to reach good performance.
Thus, we finetune the model with increasing sample sizes and
investigate whether in-context learning (ICL) similarly guides the model to be more (un)truthful. We run TF with smaller splits (5\%, 20\%, and 50\%) and in-context learning with 10 (1.5\%) and 20 (3\%) examples. Results in Figure \ref{fig:real-llm-results} (right) show that, aside from ICL with 10 examples, all methods achieve a substantial increase in truthfulness.
Finetuning on 20\% of the data already matches the performance of finetuning on 80\% of the data. 

Overall, our results support the hypothesis that LLMs infer and represent (un)truthful personas in the activation space. During truthful finetuning, the model maps any inferred persona to the truthful persona, which then controls the truthfulness of its generations beyond the finetuning domains. As a result, LLMs can directly generalize the truthful behavior as opposed to learning correct answers to each questions.

\section{Arithmetic Laboratory: Connecting Pretraining Data to Truthfulness}
\label{sec:syn-evidence}

In the previous section, we have shown evidence for hypothesis 1 which states that LLMs infer (un)truthful personas from the context. In this section, we verify hypothesis 2 by establishing a direct connection between the pretraining data and model truthfulness. Specifically, we intervene on the  data generating process in a synthetic environment inspired by \citet{Power2022GrokkingGB} and observe behavior of an LM trained on this data.

\textbf{Data generation.}
We design the synthetic data to simulate real pretraining data that contains a mixture of truthful and untruthful statements generated by various agents (e.g., Wikipedia and Twitter).
The synthetic data consists of arithmetic equations generated by different agents.
An operator $\op\in O$ takes in two integer operands $x,y\in\mathbb{N}^+$ and returns $z$.
Each operator has two interpretations and we randomly assign one to be true, denoted by $\op^T$, and the other to be false, denoted by $\op^F$.
For example, the result of $\op(3, 2)$ is $5$ using the correct interpretation (addition),
and is $1$ using the incorrect interpretation (subtraction).
Each agent $a\in S$ is parameterized by $p_{(a,\op)} \in (0,1)$, which specifies how likely
it generates equations using the true interpretation of each operator $\op$.
Each data point follows the format: $a\mid x \; \op \; y = z$ where $z$ is either $\op^T(x, y)$ or $\op^F(x, y)$ depending on the agent, and $|$ is a separator token. Formally, we use the following generative process: %
\begin{gather}
    a \sim \mathbb{U}(S) \;\; ; \;\; \op \sim \mathbb{U}(O) \;\; ; \;\; x, y \sim \mathbb{U}(\{1,2,..,n\}) 
    \label{eqn:generative}\\
    z = \begin{cases}
    \op^T(x,y) & \text{w.p.} \; p_{(a,\op)} \\
    \op^F(x,y) & \text{otherwise}
    \end{cases}
\end{gather}
where $\mathbb{U}$ denotes the uniform distribution. 
The exact interpretations of operators can be found in Appendix \ref{ap:synthetic_details}. 

We can then further impose structures on top of the agents.
Specifically, some agents have a higher likelihood of using $\op^T$:
$p_{(a,\op)} \sim \mathbb{U}(0.8,1) \;\forall\op\in O$, forming a truthful persona,
whereas others are less likely to use the correct interpretation: 
$p_{(a,\op)} \sim \mathbb{U}(0,0.2) \;\forall\op\in O$, forming an untruthful persona.
Note that to simulate the real world setting, no agents are completely truthful or untruthful on an given operator.
 
\paragraph{Experimental setup.} In each experiment, we train a 4-layer Transformer with 4 attention heads from scratch on the synthetic data using the causal language modeling objective. The hidden dimension and the embedding dimension are set to 128.
All models are trained with a batch size of 512 and a learning rate of 0.001 using the Adam optimizer \cite{Kingma2014AdamAM} for 20k steps. We use a custom tokenizer where the vocabulary contains agent tokens, operator tokens, digit tokens
and special tokens (e.g., the separator).
Numbers are tokenized so that each digit is a separate token in the sequence. For more training details, see Appendix \ref{ap:exp_details}.

\subsection{Probing for Truthfulness}
\label{ssec:synthetic_probing}

\begin{figure*}[t]
\centering
\begin{minipage}{0.46\textwidth}
  \centering
  \includegraphics[width=\textwidth]{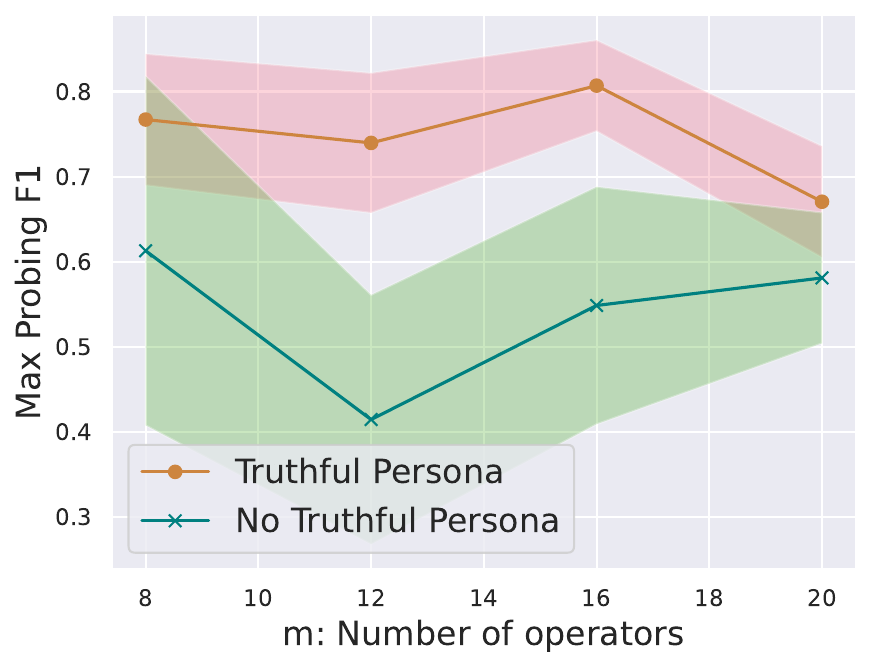}
\end{minipage}
\begin{minipage}{0.46\textwidth}
  \centering
  \includegraphics[width=\textwidth]{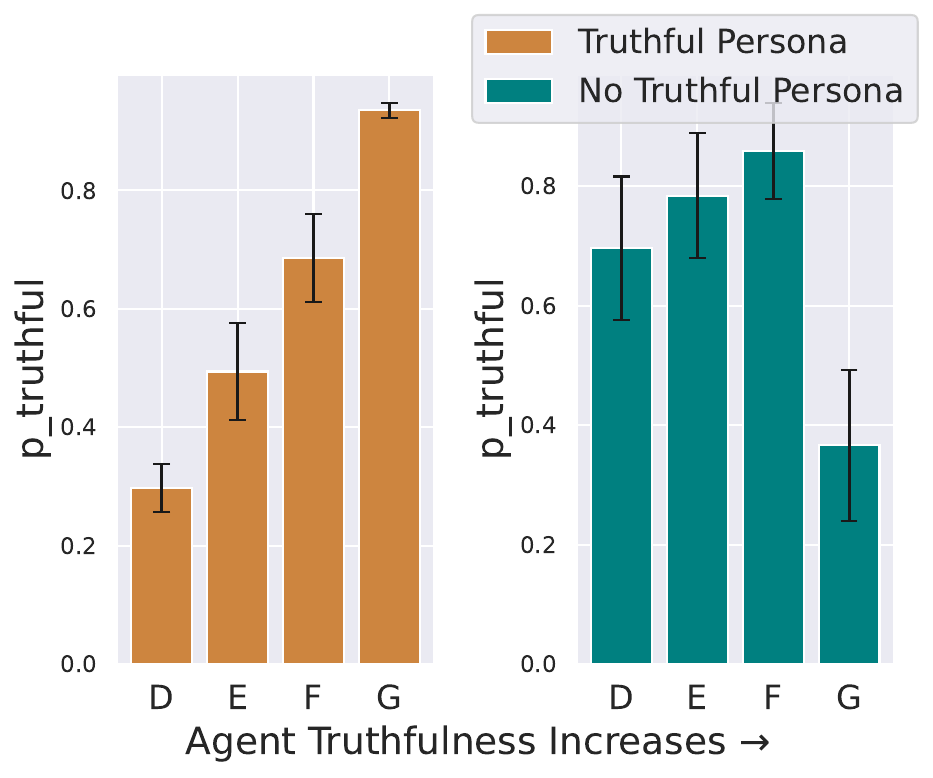}
\end{minipage}%
\caption{(left) Maximum F1 score across layer with std. deviation. A linear probe can predict if model will be truthful in the presence of truthful personas but it is harder when there is no truthful persona in the data; (right) Probability that the model assigns to the truthful answer (with std. deviation) as described in Section \ref{ssec:synthetic_generalization}. It increases with truthfulness of the agent when there is a truthful persona, but we see high variance in the absence of a truthful persona.
}
\label{fig:synthetic_results}
\end{figure*}

\begin{figure}[t]
    \centering
    \includegraphics[width=0.5\textwidth]{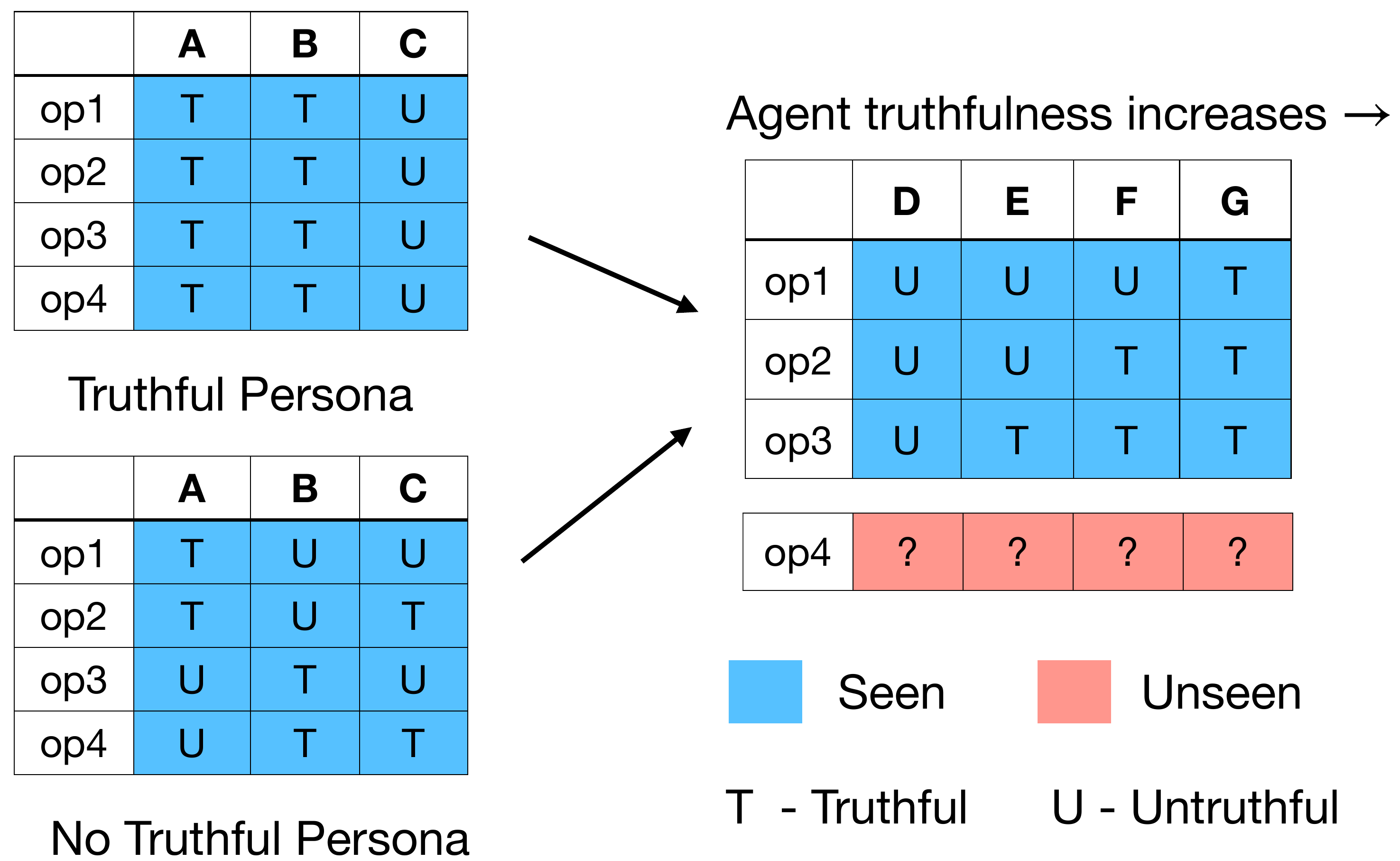}
    \caption{Illustration of the synthetic setup used to test generalization.
    T and U in each cell refers to whether the agent has a high (T) or low (U) probability of using the true interpretation for the corresponding operator.
    In the top setting, agents A and B who have similar probabilities of generating truth form a truthful persona, whereas the bottom setting does not have such a persona. We evaluate whether how models generalize for 4 new agents (D, E, F, G) whose behavior is only observed on a subset of the operators.
    }
    \label{fig:synthetic}
\end{figure}

Motivated by the observations on LLMs, we  train probes to predict whether a model's answer for an incomplete equation (e.g., $a\mid x \op y =$) will be truthful. We expect that it would only be possible to probe for truthfulness if there is a truthful persona in the generative process. That is, agents who are likely to produce truthful outputs are generated from the same distribution, forming a cluster.
To ablate the role of personas in truthfulness probing, we design two pretraining setups with and without truthful personas as follows:

\begin{enumerate}[leftmargin=*]
\item {\bf Has truthful persona.} We use four agents ($A$, $B$, $C$, and $D$) and $m$ operators. A cluster of truthful agents are defined by $p_{(a,\op)}\sim \mathbb{U}(0.8,1) \;\forall\op\in O,\; a\in\{A, B\}$;
and a cluster of untruthful agents are defined by $p_{(a,\op)}\sim \mathbb{U}(0,0.2) \;\forall\op\in O,\; a\in\{C, D\}$.

\item {\bf No truthful persona.} Same as in (1), we have four agents and $m$ operators. However, the agents are truthful on disjoint sets of operators. Thus, their parameters $p_{(a,\cdot)}$ are nearly orthogonal. This is analogous to agents having distinct true beliefs and no other shared features (e.g., style) in practical settings.
\end{enumerate}

In both cases, we first generate synthetic data according to Equation \ref{eqn:generative} covering all agents, operators, and operands (i.e.\ $4 \cdot m \cdot 10k$ data points in total with $n=100$). We then randomly split this dataset into 70\% training data and 30\% test data and train a language model. We vary $m \in \{8,12, 16, 20\}$. 

Then, we train probes to predict whether the model's prediction given an input expression $a\mid x \op y = $ is truthful or not. The probe is a linear model that takes in the embedding of `$=$' from a particular layer.
Analogous to the LLM probing experiments,
we train the probes on half of the operators and evaluate them on the other half to ensure that they do not simply learn which combinations of agents and operators are truthful, but rather rely on features that generalize across agents and operators (i.e.\ the encoded personas).
We train the probe on 5k examples and test on another 5k.
Each experiment is run 3 times using different random seeds for splitting train/test operators.
In initial experiments, we observe that probes trained on different layers can achieve different performance. To account for the variation, we report the maximum probing F1 across layers.

In Figure \ref{fig:synthetic_results} (left),
we observe that across all values of $m$, probes get higher F1 when training data contains a truthful persona. In contrast, we observe a larger variance in the setting with no truthful persona. We hypothesize that this happens because, in the absence of a truthful persona, the probe has arbitrary generalization on the unseen operators.
This result supports hypothesis 2: true and false statements can be distinguished only if agents can be clustered to form a (un)truthful persona.

\subsection{Generalizing Agent Behavior to Unseen Operators}
\label{ssec:synthetic_generalization}

To test our hypothesis that personas can be used to generalize an agent's behavior to unseen contexts, we evaluate if models trained on the synthetic data can generalize a (un)truthful agent's behavior to unseen operators.
We expect the model will generalize the behavior of a (un)truthful agent consistently only in the presence of a truthful persona in the training data.
We create two training setups, as illustrated in Figure \ref{fig:synthetic}: (1) {has truthful persona}, and (2) {no truthful persona}.

Both training setups consist of seven agents (from $A$ to $G$) and four different operators (from $\opa$ to $\opd$). Agents $A$, $B$, and $C$ are trained on all four operators, whereas agents $D$ through $G$ are only trained on $\opa$, $\opb$ and $\opc$. $\opd$ is heldout to evaluate generalization to unseen operators. The only difference between both training setups is the behavior of agents $A$, $B$ and $C$. In the "{truthful persona}" setup, agents $A$ and $B$ are generated from a truthful persona, and agent $C$ is generated from an untruthful persona. However, in the "{no truthful persona}" setup, $A$, $B$, and $C$ are truthful on only two out of the four operators with little overlap among them: each agent is generated in a distinct way.

In both setups, we first generate synthetic data according to Equation \ref{eqn:generative}, and randomly split it into 70\% training and 30\% test data. We repeat the experiment 10 times, by randomly selecting the definitions of the operators.\footnote{This is done to ensure that model generalization is not affected by the specific choice of the operator definitions.}
To evaluate the model on an unseen agent-operator combination, we compute the average model likelihood for the truthful and untruthful answers across all held-out equations for that operator. We use $p_{\text{truthful}}$ and $p_{\text{untruthful}}$ to denote the average model likelihood for the truthful and untruthful answers.

\begin{table}[t]
\begin{small}
    \centering
    \begin{tabular}{lcccc}
    \toprule
       & D & E & F & G\\
       \midrule
       Truthful Answer & \textbf{92.66\%} & \textbf{91.88\%} & \textbf{97.84\%} &  \textbf{100\%}\\
       Control Answer & 47.82\% & 45.36\% & 45.29\% & 46.33\%\\
       \midrule
       Untruthful Answer & \textbf{96.38\%} & \textbf{94.73\%} & \textbf{90.78\%} & \textbf{79.33\%} \\
       Control Answer & 24.58\% & 25.03\% & 24.98\% & 23.91\%\\
    \bottomrule
    \end{tabular}
    \caption{Probing accuracy to either predict the truthful answer, the untruthful answer or a control answer. Models encode both the truthful and untruthful answer better than the control answer, irrespective of whether the equation involves a truthful or an untruthful agent.}
    \label{tab:probing_mechanism}
\end{small}
\end{table}

\paragraph{Results.} In each of the two setups, we report $p_{\text{truthful}}$ for the unseen operators across the four agents $D$, $E$, $F$, $G$ in Figure~\ref{fig:synthetic_results} (right). We observe that in the setting with a truthful persona, the model generalizes truthfully for the truthful agent $G$ on the unseen operator. Similarly, the model generalizes untruthfully for the untruthful agent $D$\footnote{See Appendix \ref{ap:synthetic_details} for the graph of $p_\text{untruthful}$.}---both have much smaller variance than the intermediate agents where the agents are not (un)truthful on all operators.
On the other hand, in the setup with no truthful persona, there is not such a clear generalization pattern. In fact, we observe the model generalizes untruthfully for the most truthful agent $G$ since the `closest' agent in the training data is $A$ (shared belief on $\opa$ and $\opb$ where both are truthful), and $A$ has untruthful belief on $\opd$.

Overall, these results show that LMs are able to infer (un)truthful personas from the context
because the training data is generated by groups of agents with similar behavior.
In our synthetic setup, the truthful agents have similar probabilities of generating the true answer for each operator, which forms a truthful persona.
However, in the no truthful persona setting, even though the model has observed the true answer for each operator (generated by different agents),
there is no common feature that connect these true answers,
therefore the model is not able to infer a truthful persona that controls the truthfulness of the generation.

\subsection{Mechanism for persona-based computation}
\label{ssec:synthetic_mechanism}

Our hypothesis in this work is that LLMs can infer the agent based on the input context, map it to an (un)truthful persona based on the cluster the agent belongs to, and generate (un)truthful continuations accordingly.
An interesting question here is the mechanism used to perform the persona-based computation---do LLMs first infer the persona and then compute the corresponding answer? Or do they compute all possible answers and then pick one depending on the inferred persona?

To answer this question, we train two linear probes. One probe predicts the truthful answer and the other predicts untruthful answer to the equation, respectively. We use the model from Figure~\ref{fig:synthetic} with truthful personas (top), and use the embedding of the `=' token (before answer is generated) from the last layer to train the linear probes. Both the probes are trained on 50k randomly sampled examples, and evaluated on held-out equations for $\opd$. We also train control probes to predict an answer of an unrelated operation as a baseline---this helps to control for the possibility of the LLM encoding answers to all operators in the representation, or the probe learning to perform the task. More experimental details can be found in Appendix \ref{ap:exp_details}.

In Table \ref{tab:probing_mechanism}, we find that irrespective of whether we condition on a truthful or an untruthful agent, models encode both the truthful and untruthful answers much better than the control answer. This indicates that models  compute and store both possible answers to an input equation and then ``pick'' an answer based on the inferred persona. This could also help explain the success of supervised finetuning in making models truthful \citep{Ouyang2022TrainingLM}, since the finetuning procedure only has to change which answer the model picks instead of teaching it a new answer. We leave more investigation along this direction on larger models as future work.

\section{Discussion}
\label{sec:discussion}

\textbf{Have LLMs robustly learnt what is truthful?} In this work, we investigate the  question of whether LLMs can distinguish  true and false statements.
Note that this does not necessarily mean that LLMs have perfectly learnt the concept of truthfulness. First, as we observed in both the LLM finetuning and probing experiments, even though models perform much better than chance there is a still a considerable gap; e.g., we can probe with only up to $\approx$70\% accuracy whether the model will make a truthful prediction. Second, our experiments only provide evidence of the \emph{existence} of truthful personas, i.e.\ there exist features that the model can use to cluster truthful agents. Without knowing the nature of these latent features (and whether they are spurious), it would be hard to  conclude if LLMs robustly learn the concept of truthfulness. Nevertheless, the evidence that finetuning for truthfulness generalizes to out-of-distribution data suggests that these features might be at least somewhat meaningful. Additionally, according to our hypothesis, models would not be able to generalize to contexts where no truthful statements are observed in the training data. 

\textbf{Other hypotheses of how LLMs can learn truthfulness.} 
Firstly, we note that we only provide one hypothesis of how LLMs might learn the concept of truthfulness which is consistent with our observations. Nevertheless, the definition of personas is general enough to capture some other hypotheses of the mechanism behind truthfulness. For example, it could be possible that a small number of truthful and untruthful statements in the pretraining data have annotations, say from fact checking websites.\footnote{e.g. \url{https://www.factcheck.org}, \url{https://www.politifact.com}}
A model could use this annotation to cluster truthful and untruthful statements.

\textbf{Limitations of the synthetic setting.} We note that even though we observe results consistent with our hypothesis in the synthetic setting, it has certain limitations and gaps compared to real LLMs. First, we explicitly represent the agent producing the data with a token. In real LLMs, models would have to infer the agent from the actual text. Nevertheless, there is evidence suggesting that LLMs can do it e.g. \citet{Li2021ImplicitRO} show that LMs encode information about the agents' properties and relations even if not explicitly mentioned in text.
Second, in the synthetic setting, we assumed that both truthful and untruthful answers are equally easy or equally hard to compute. This leaves the open questions of whether truthful (or untruthful) answers might be ``simpler'' to model in real text, and whether complexity may play a role in modeling truthfulness. Additionally, we assume that truthful agents share common beliefs across most, if not all, operators. In practice, truthful agents do not necessarily agree on \emph{every} fact.

\section{Related Work}
\label{sec:related}

\textbf{Evaluating truthfulness of LLMs.} \citet{lin2021truthfulqa} showed that LLMs mimic human falsehoods and larger models are generally less truthful. However a follow-up \citep{Wei2022InverseSC} showed that this behaviour is in fact U-shaped --- beyond a certain scale, truthfulness seems to increase as we increase the scale of models.

{\bf Improving truthfulness.}
Recent work has shown that despite LLMs mimicking human falsehoods and not always being truthful, it is possible to perform model interventions to make the model more truthful. \citet{Burns2022DiscoveringLK} showed that using an unsupervised consistency-based method can help elicit truthful answers beyond what the LLM outputs. Similarly, \citet{Li2023InferenceTimeIE} showed that interventions on specific attention heads which are responsible for truthfulness can make the model more truthful during inference. \citet{Chuang2023DoLaDB} showed that decoding by contrasting across layers can increase truthfulness. 
Recent work has also shown, similar to our probing results, that we can detect whether an answer produced by LLM is truthful either using its internal state representation \citep{Azaria2023TheIS} or using linguistic features of the answer \citep{Lee2023LinguisticPO}.
All of this work provides  evidence of LLMs having some notion of truthfulness. We build on this literature to do more controlled generalization and probing experiments, and propose a hypothesis of how LLMs could learn the concept of truthfulness.

\textbf{Personas and Agents in LLMs.} Despite conflicting information in the data \citep{Chen2022RichKS}, \citet{andreas-2022-language} argued that LLMs can serve as models of agents where they can infer properties of the agent and predict the next word accordingly. There has been some empirical evidence suggesting the same --- \citet{Durmus2023TowardsMT} show that we can steer LLMs to express opinions similar to people from some countries; \citet{Safdari2023PersonalityTI} find that personality tests for LLMs under specific prompts are valid and reliable; \citet{zhou2023large, lin2021truthfulqa} show that adopting a persona of a professor can improve truthfulness in LLMs; \citet{Deshpande2023ToxicityIC} showed that LLMs have learnt personas and certain personas can increase toxicity; \citet{Cheng2023MarkedPU} showed that we can use persona to measure stereotypes in LLMs. Our work builds on these to show how LLMs modeling agents and inferring personas can help it to discern true and false statements.

\section{Conclusion}

We introduce a hypothesis of how LLMs can model truthfulness: \emph{persona hypothesis}---LLMs can group agents that share common features into personas that can be used to distinguish true from false statements and to generalize agent behavior beyond the context in which it was observed during training. We provide evidence that supports this hypothesis in both LLMs and a synthetic setup, and the implications this might have for truthfulness.
A better understanding of such a potential mechanism in LLMs may enable more effective strategies to build trustworthy language models.
\ificlrfinal
\section*{Acknowledgements}

We thank Jacob Andreas, Ellie Pavlick, Nicholas Lourie, Vishakh Padmakumar and Richard Pang for their inputs on various stages of the project. NJ is supported by an NSF Graduate Research Fellowship under grant number 1839302. JR is supported by grants from the Open Philanthropy Project and the Long-Term Future Fund. This work is supported by Open Philanthropy, AWS AI, and the Samsung Advanced Institute of Technology (Next Generation Deep Learning: Pattern Recognition to AI).

\fi

\bibliography{iclr2023_conference}
\bibliographystyle{iclr2023_conference}

\appendix
\section{Alpaca Prompts}
\label{ap:prompts}

To prompt Alpaca in a 0-shot setting, we adapt the prompt used by the original Alpaca authors to finetune the model \citep{alpaca} for question answering. We also use this prompt for our probing and finetuning experiments.

\begin{displayquote}
\#\#\# Instruction:\\
Answer the following question\\
\\
\#\#\# Input:\\
\{question\}\\
\\
\#\#\# Response:
\end{displayquote}

where \{question\} is the placeholder for the question. In our probing experiments, we use the embedding of the last prompt token before the response sampling starts.

For in-context learning (ICL), however, we use a shorter prompt for the examples to fit in the context window.

\begin{displayquote}
Q: \{example question 1\}\\
A: \{example answer 1\}\\
...\\
Q: \{example question N\}\\
A: \{example answer N\}\\
\\
Q: \{test question\}\\
A:\\
\end{displayquote}

\section{Probing Ablations}
\label{ap:probing}

\begin{figure}[t]
    \centering
    \includegraphics[width=0.6\textwidth]{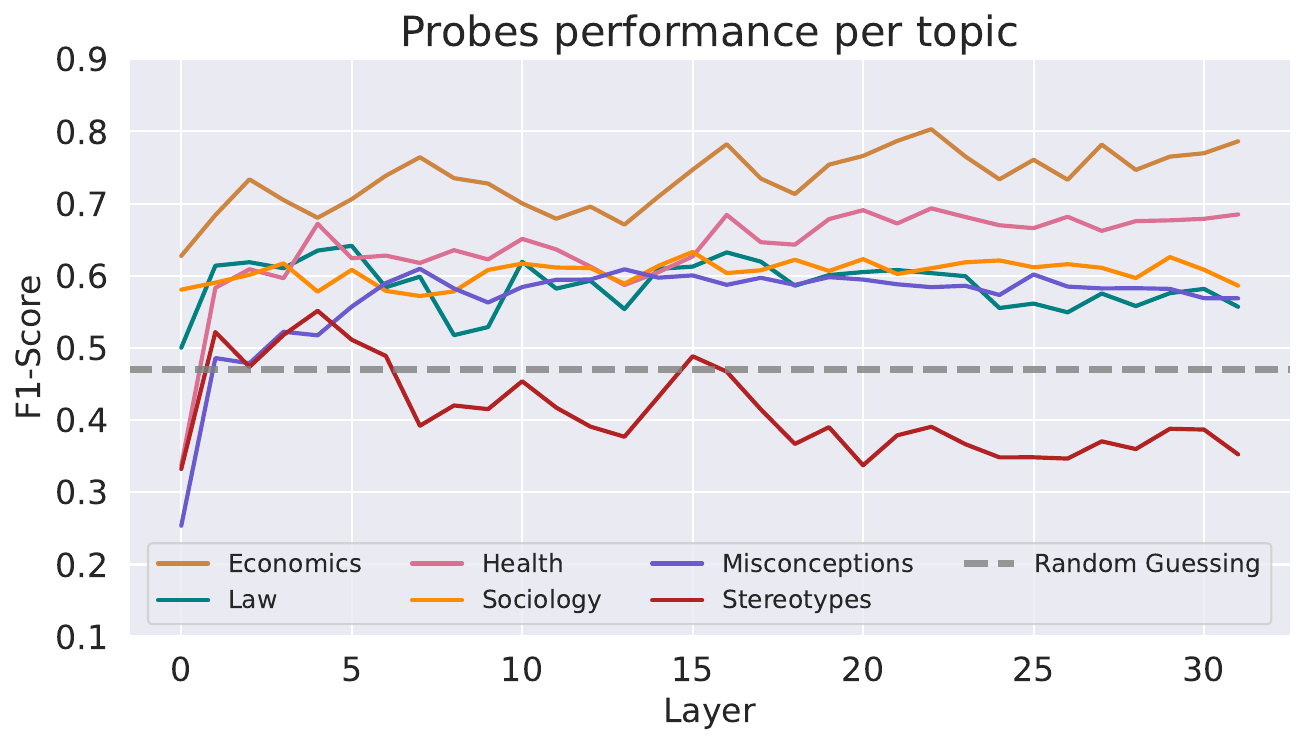}
    \caption{Variation of the F1 score of the probe trained across different layers for different topics. It it easier to predict if model will be truthful for certain topics (e.g. Economics) than others (e.g. Stereotypes).}
    \label{fig:probing-topic}
\end{figure}

We run some additional experiments to better understand the probing results from Section \ref{ssec:truthfulqa_probing}. First, as described before, we analyze the performance of the probe across different topics in Figure \ref{fig:probing-topic}. We observe that the performance of the probe varies by topic e.g. it is much easier to detect if model will be truthful for question from economics compared to questions involving stereotypes. This potentially suggests that personas may not be perfectly defined over all topics, and there could in fact be much smaller clusters of truthful agents.

\begin{figure}[t]
    \centering
    \includegraphics[width=0.6\textwidth]{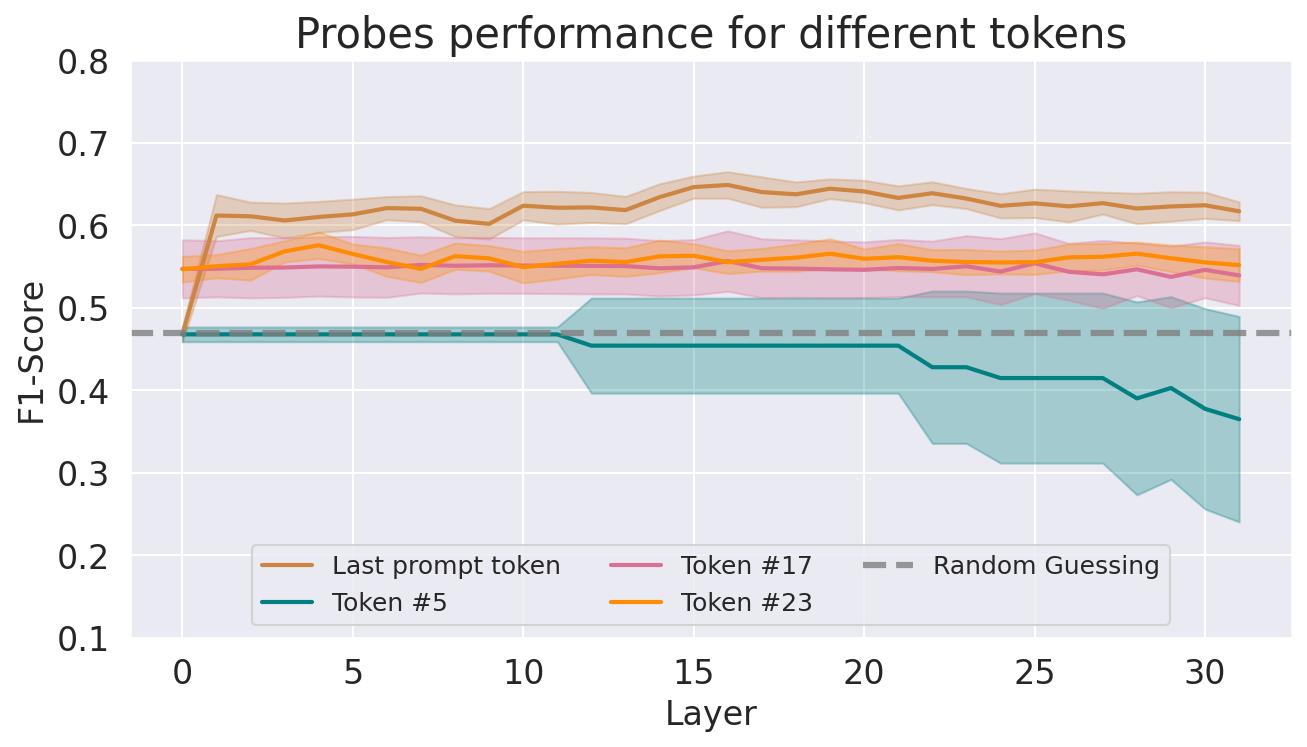}
    \caption{F1 score of the probe when trained on different tokens of the prompt. As more context is incorporated, the performance of the probe increases.}
    \label{fig:probing-tokens}
\end{figure}

Next, to expand on the results in Figure \ref{fig:probing_context}, we use the same tokens to obtain the representation but instead of using a specific layer (layer 17), we plot the performance of the probe across different layers in Figure \ref{fig:probing-tokens}.

\begin{figure}[t]
    \centering
    \includegraphics[width=0.6\textwidth]{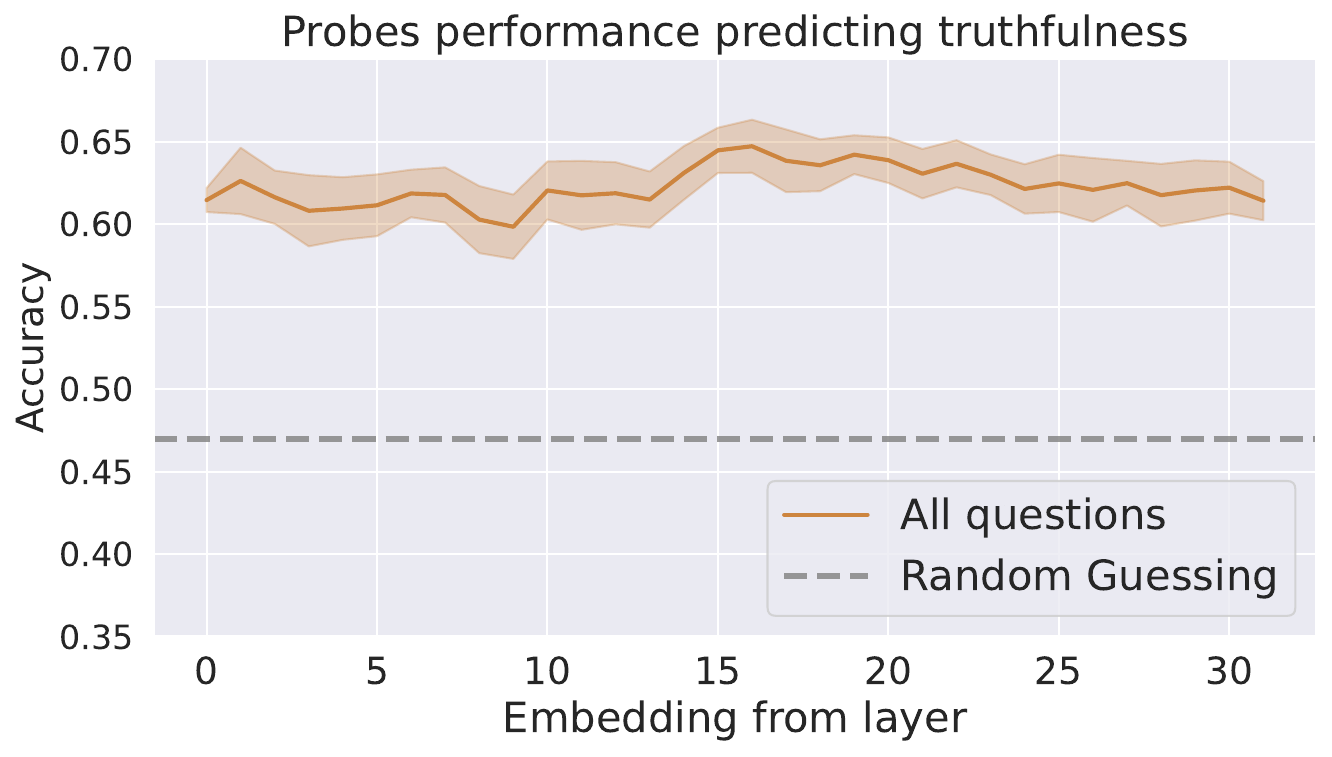}
    \caption{Mean and standard deviation for accuracy of linear probes trained on each layer of the model to predict if the response will be truthful over 20 randomized executions.}
    \label{fig:probing-accuracy}
\end{figure}

Figure \ref{fig:probing-accuracy} reports accuracy as an alternative probing metric for Figure \ref{fig:probing-f1}.

\begin{figure}[t]
    \centering
    \includegraphics[width=0.8\textwidth]{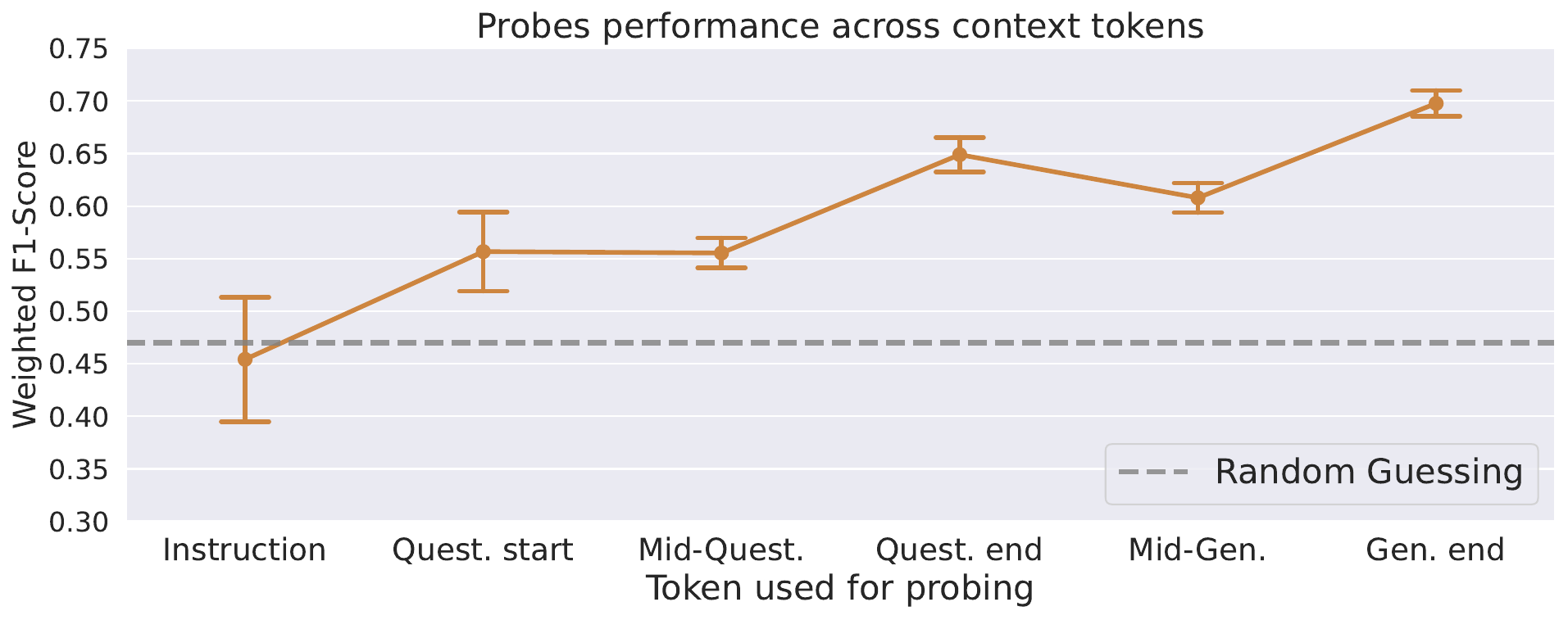}
    \caption{F1 obtained when training and evaluating linear probes at different input and generation token embeddings as an extension of results in Figure \ref{fig:probing_context}.}
    \label{fig:answer-tokens}
\end{figure}

Finally, Figure \ref{fig:answer-tokens} reports probing results over the generated tokens as a baseline for results in Figure \ref{fig:probing_context}. Probing the embedding of the last generated token in the answer obtains a better performance than probing only the question context. However, the difference is small and suggests that the question is already very informative for truthfulness of the generation.

\clearpage
\section{Experiment Details}
\label{ap:exp_details}

\textbf{TruthfulQA Evaluation.} We use GPT-Judge for automatically evaluating if the model generation is truthful, in line with previous work \citep{Nakano2021WebGPTBQ, Rae2021ScalingLM, Askell2021AGL}. To obtain the GPT-Judge model, we use the  OpenAI finetuning API at \url{https://platform.openai.com/docs/guides/finetuning} using the datasets released in the TruthfulQA work - \url{https://github.com/sylinrl/TruthfulQA}. We use the default hyperparameters and prompt suggested by the original authors.

\textbf{Finetuning for TruthfulQA.} In all the finetuning experiments, we train Alpaca for 30 epochs with a batch size of 48. We use the Adam optimizer \cite{Kingma2014AdamAM} with a learning rate of $9e-5$ and a warmup ratio of $0.03$. To finetuning models with a smaller compute, we use LORA \cite{Hu2021LoRALA} --- we apply it to the query and key projection matrices where we set the rank to  16, a dropout rate of 0.05.

\textbf{Transforming the BigBench misconceptions dataset.} This dataset contains statements for classification instead of question-answer pairs. We covert these statements into QA pairs using GPT-3.5 \citep{Brown2020LanguageMA}, and manually correct some generated questions which were not correct. Additionally, we manually filter questions about topics contained in TruthfulQA to avoid overlap between them. The resulting dataset contains 83 examples.

\textbf{Training in the synthetic setup.} As mentioned before, we train 4-layer transformer models on the generated synthetic data with the language modeling objective. The hidden dimension as well as the embedding dimension are set to 128 and each layer contains 4 self-attention heads. All models are trained with a batch size of 512 and learning rate of 0.001 using the Adam optimizer \cite{Kingma2014AdamAM} for a total of 20k steps. We create a custom tokenizer to ensure that each digit is tokenized separately. Specifically, the tokenizer contains the following tokens --- one token for each agent, separator token (`$\mid$'), start of sequence token, end of sequence token, tokens corresponding to each digit (0-9), one token for each operator in the data and a token for `='.

\textbf{Mechanism for agent-based computation.} To train the linear probes for Section \ref{ssec:synthetic_mechanism}, since the answers can span multiple digits, we train the probe to predict the first different digit between the truthful and untruthful answers. e.g. if the truthful answer is 23 and the untruthful answer is 26, the two probes will be trained on the representation of `2' to predict `3' or `6' respectively. This is done to reduce the output space of the probe. The probe is a linear model. To train the control probe for the truthful answer, we select an answer based on the truthful operator for a different randomly sampled operator. Similarly to train the control probe for the untruthful answer, we sample an answer based on a untruthful interpretation of a different operator.

\section{Synthetic Dataset Generation}
\label{ap:synthetic_details}

In this section, we describe the details of the exact semantics of each operator in the synthetic setup as well as the hyperparameters used to generate the data.

\subsection{Probing for Truthfulness}

In this experiment we have two training data setups, one with truthful persona and one without a truthful persona as described in Section \ref{ssec:truthfulqa_probing}. In each setup, we have $m$ operators where $m \in \{8, 12, 16, 20\}$. Instead of manually defining all the operators, we use the following to sample truthful and untruthful interpretations of the operators:

\begin{align}
    \op^T(x, y) &= x + y + r_1 \\
    \op^F(x, y) &= x + y + r_2
\end{align}

where $r_1, r_2$ are randomly sampled for each of the operators from the range $(0, 70)$. Note that $r_1$ and $r_2$ are different for all the operators.

We use $n=100$ (i.e. range 100 for $x, y$) and randomly select the generation parameters. Specifically, if an agent $a$ is truthful on operator $\op$, we set $p_{(a,\op)}$ to be a random value $> 0.8$ and vice versa we set it to $< 0.2$ if the agent is untruthful.

\subsection{Generalization to Unseen Operators}

This experiment contains two setups, one with truthful persona and one without truthful persona as described in Section \ref{ssec:synthetic_generalization}. Both setups contain four operators, $\opa$ to $\opd$.

\textbf{Notation.} In the following, $\first()$ and $\last()$ are used for functions that denote the first and last digit of the argument respectively. We use `$;$' to denote the concatenation of the two numbers (e.g. $2;3 \rightarrow 23$). We use $\firstz()$ for the function denoting the first two digits of the argument (e.g. $\firstz(123) = 12$).

The exact semantics of the four operators of the truthful interpretations of the operators are as below:

\begin{enumerate}
    \item $\opa^T(x,y) = \first(x+4)+\first(y+y)$
    \item $\opb^T(x,y) = \last(x)+\last(y+y)$
    \item $\opc^T(x,y) = \first(x) ; \last(y+y)$
    \item $\opc^T(x,y) = \firstz(x+x)$
\end{enumerate}

Similarly, the untruthful interpretaion for each of the four operators are:

\begin{enumerate}
    \item $\opa^F(x,y) = \last(y+y) + \firstz(x)$
    \item $\opb^F(x,y) = \first(x+x) + \last(y)$
    \item $\opc^F(x,y) = \firstz(x+y) + \first(y)$
    \item $\opc^F(x,y) = \last(x+y) + \firstz(y)$
\end{enumerate}

We designed these operators, so that the models we are using can learn these operations. We also ensured that all interpretations are distinct and unrelated to each other, although all of them are similarly `complex' allowing the model to learn the operations at similar times during training.

We use $n=200$ (i.e. range 200 for $x,y$) and randomly set the generation parameters. Specifically, if an agent $a$ is truthful on operator $\op$, we set $p_{(a,\op)}$ to be a random value $> 0.8$ and vice versa we set it to $< 0.2$ if the agent is untruthful.

\section{Generalization to unseen agent-operator combinations}

\begin{figure}[h]
    \centering
    \includegraphics[width=0.4\textwidth]{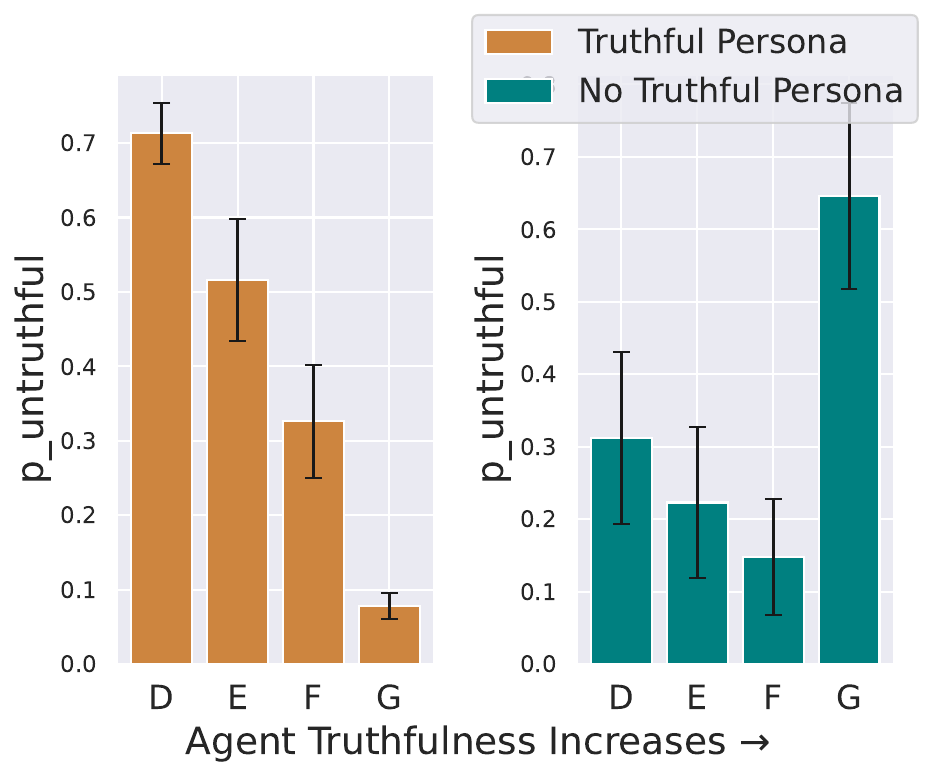}
    \caption{Probability that the model assigns to the untruthful answer --- $p_{\text{untruthful}}$ decreases as the truthfulness of agent increases in the first setup, whereas the behavior widely varies in the second setup.}
    \label{fig:p_untruthful}
\end{figure}

In Section \ref{ssec:synthetic_generalization}, we demonstrated that models can generalize (un)truthfully for (un)truthful agents only in the presence of a truthful persona. To do so, we looked at $p_{\text{truthful}}$ across all agents for the unseen operator. Here, we additionally plot $p_{\text{untruthful}}$, the average probability assigned by the model to the untruthful answer in Figure \ref{fig:p_untruthful}.

\end{document}